# Neurofeedback-Driven 6-DOF Robotic Arm: Integration of Brain-Computer Interface with Arduino for Advanced Control

Ihab A. Satam[1], Róbert Szabolcsi[2]

[1]Doctoral School on Safety and Security Sciences, Óbuda University, Hungary
[1]Mechatronics Department, Technical College-Alhawija, Northern Technical University, Iraq
[2]Department of Measurement Technology and Automation, Institute of Electronic and Communication Systems, Kandó Kálmán Faculty of Electrical Engineering, Óbuda University, Hungary
ihab.satam@uni-obuda.hu, szabolcsi.robert@kvk.uni-obuda.hu

## Abstract

Brain computer interface (BCI) applications in robotics are becoming more famous and famous. People with disabilities are facing a real-time problem of doing simple activities such as grasping, handshaking etc. in order to aid with this problem, the use of brain signals to control actuators is showing a great importance. The Emotive Insight, a Brain-Computer Interface (BCI) device, is utilized in this project to collect brain signals and transform them into commands for controlling a robotic arm using an Arduino controller. The Emotive Insight captures brain signals, which are subsequently analyzed using Emotive software and connected with Arduino code. The HITI Brain software integrates these devices, allowing for smooth communication between brain activity and the robotic arm. This system demonstrates how brain impulses may be utilized to control external devices directly. The results showed that the system is applicable efficiently to robotic arms and also for prosthetic arms with Multi Degree of Freedom. In addition to that, the system can be used for other actuators such as bikes, mobile robots, wheelchairs etc.

## Introduction

The loss of limbs causes a tremendous effect on the injured person in daily life activities. People with upper limbs loss face challenges in activities like throwing a ball, driving a car or handshakes etc. The idea of controlling actuators such as robotic / prosthetic arms, mobile robots, cars, bikes etc. [1], [2], [3], [4], [5], [6], [7] Using brain signal has developed and improved in the last decades. However, there is a lot to investigate yet. The method contains several steps. First, is to collect the signals non-invasively from the subject brain [8]. Secondly, is to preprocess these signals and remove all the artifacts and noise to prepare them for the next step, the classification. After that we can use these signals to convert it to command to move the actuators, depending on the applications they are used in. The BCI (Brain Computer Interface) enables communications between the brain and machines during the recording of electrical activity and the connection of external devices. BCI is a device that can detect and collect the EEG signals produced by the brain and convert them into commands to perform a desired action for an external device connected to it [9]. The non-invasive device is cost effective with high potential that can be used for different human-robot applications such as Wheelchairs, prosthetics etc. The idea of BCI is to collect the signals from the scalp; the desired collected information is converted to an active function. The microcontroller is operated by this active function [10]. Based on the input signals the microcontroller determines the DOF and the robotic arm acts according to the received signal. The robot kinematic model involves the robot`s motion regardless of the forces that produce the motion [11]. Robot kinematics studies how one link in the arm is moving with respect to one another and with time, it deals with the geometric and time properties of the motion. It also studies the relationship between the position and orientation of robot end effector and its joints. There are two types of robots kinematic:

*1- Forward Kinematic*

It is the determination of an end effector's position and orientation utilizing information from the joints and linkages that connect to it. Forward kinematics may be used to compute the position and orientation of the end effector from the present locations, angles, and orientation of the joints and linkages [12]. While correct positioning and orientation of the end effector is critical, it is dependent on a thorough understanding of the variables of the joints and linkages that connect to the end effector, as well as a solvable model.

*2- Inverse Kinematic*

It is the process of calculating the variables associated with a collection of joints, and linkages that coupled to an end effector. Given the position and orientation of the end effector, inverse kinematics may be used to determine the variables associated with those joints and linkages, such as position, angle, and orientation. IK uses Jacobian matrix to compute the optimal path between positions [13]. A 6 Degree of Freedom DOF robotic arm is widely used in several applications, especially in factories. The 6 DOF robotic manipulator is also the closest thing to a human arm as will be explained well in the methodology section. The microcontroller has an important role in controlling physical parts of the system [14]. It has the capability of handling multiple inputs and outputs at the same time, which makes it ideal for motor controlling. Arduino controller is an open-source platform, it has the ability to read inputs from different sources, process it and convert it to an output to control an external actuator. Over the last decade, there have been considerable advances in the use of brain-computer interfaces (BCIs) to operate robotic systems.

## Preliminary Studies

Several studies have been implemented in this field. In regard to EEG signal recording and processing, Rashmi et al. [15] developed a MATLAB application that helps with mental stress healing. This approach has three modules to work with, the first is offline EEG signal recording and analysis, the second module is mental stress healing, and the last module is real-time EEG data analysis. Arman and his team [16] used a BCI system in speech prediction application to aid people with speech disorder. The system designed to recognize 11 Bengali vowels with 84,02% of accuracy.

The team in [17] proposed a multimodal approach to evaluate the state of wakefulness in users during a BCI task using SSVEP. The study showed that the low state has a negative influence on the BCI performance. The study of Rimbert et al. [18] examined how long-term kinesthetic motor imagery (KMI) exercise affected motor cortex oscillations and brain-computer interface performance, taking into account the subject's mental and environmental contexts. As a result of this study, the output of long-term KMI is continuous ERD that is not affecting the person's well-being. Most of the BCI systems in cognitive training produce the output according to output of the classifiers. The scientists in [19] demonstrated a system that allows the users to regulate the theta and beta waves over a specific region of the brain. They built a multi-channel EEG Based BCI for attention training. To classify EEG signals, there are several Machine learning algorithms. Supervised, unsupervised and reinforcement learning [20], [21], [22]. In order to classify the SSVEP signals, the team in [23] designed a new innovative classification algorithm using Machine learning and Statistical optimization with accuracy of 97.7%. Deep learning -based transformer methods show great potential in analyzing data. Mahsa et al. [24] presented a new model based on transformer to extract temporal and spectral feature from EEG signals to use them in classification process. The purpose of preprocessing and classifying the brain signals is to apply it in real applications that assist people with their daily activity. Paralyzed cases is continue to spread widely in the world, causing negative affect on people. The BCI features can assist with solving the problem by providing aid for people to control wheelchairs. The scholars in [25] designed a system that controls wheelchairs based on brain sensory activities. Another approach for controlling wheelchair designed by [26]they presented a smart rolling approach that has two different modes to control wheelchairs. First, by using the head motion only and second, using the head and hands motion, depending on the level of disability. The motor execution tasks were based on eye blink, jaw clench and hands open/close. Another application that can use the features of EEG signals is the UAV. That what Tao and his team did in [27]. In their study, they developed a modular multi-quadcopter system, followed by a 3D VR scene system with a digital twin feature for visual stimulation. A BCI system based on the Stable State Visual Evoked Potential (SSVEP) paradigm was used to control the quadcopter swarm. Subjects effectively manage multi-quadcopter formation using the proposed VR-based BCI interactive system, achieving 90% accuracy and high information transfer rates. The immersive VR twin system, developed one-to-one for EEG signal collecting, enhances participants' experience. Using the EEG application in limbs rehabilitation has shown a great deal of improvement. The Robotic exoskeleton is a piece of equipment with actuators that connect to and around the human body to help with movement by providing mechanical power [28]. The device is very beneficial especially for people suffering from spinal injury. The team in [29] presented a work that studies the use of a brain-computer interface (BCI) based on motor imagery (MI) for the control of a lower limb exoskeleton to help in motor recovery after a neurological damage. Ten healthy participants and two patients with spinal cord damage underwent BCI evaluations. Virtual reality (VR) training was used to expedite BCI training for five participants who were able-bodied. When the results from this group were compared to those from a control group consisting of five participants with normal physical capabilities, it was discovered that using VR for shorter training sessions did not lessen the BCI's effectiveness—in fact, in certain situations, it enhanced it. Patients were pleased with the method and showed the ability to manage trial sessions without becoming agitated. Brain-Machine Interfaces (BMIs) that utilize a motor imagining paradigm offer a natural method for controlling the exoskeleton while walking. However, because to accuracy limits and sensitivity to false activations, their clinical usefulness continues to be challenging. Soriano et al. [30] suggested a solution that combines the BMI with techniques based on identifying Error Related Potentials (ErrP) to self-correct erroneous commands and boost the system's usability and accuracy. In order to lessen false starts in the BMI system, the project aims to describe the ErrP at the onset of walking using a lower limb exoskeleton. Additionally, their research helped identify whether kind of feedback—visual, tactile, or visuo-tactile—performs best in evoking and identifying the ErrP. Wang et al. [31] construct a (BCI) system based on augmented reality that can be used with an exoskeleton for rehabilitation. The technology uses sequential logic decoding to control sixteen rehabilitative motions and builds a four-category BCI using HoloLens. The effectiveness of computer screen-based brain-computer interface (CS-BCI) and AR-BCI is compared in offline studies, and the effects of data length, electrode number, and electrode position on BCI performance are examined. Next, in online rehabilitation training, the movement accuracy of the exoskeleton and the instruction categorization accuracy of the AR-BCI are assessed. In offline studies, AR-BCI's average recognition accuracy was 90.2%, which is less than that of CS-BCI. When just Oz and O2 electrodes are utilized, the identification accuracy is still above 90%. The idea of hybrid BCI instructs that the combination of EEG and EMG signals can be of use to control actuators such as robotic arms. The team in [32] presented this idea in their work. They used EMG signals collected from the hand to control the motion of the robotic arms while the motor imagery of left and right hand is used to grasp command. One of the most remarkable uses of a brain-computer interface (BCI) is the control of a robotic arm. The work of [33] proposes a unique approach for BCI motor imaging (MI) electroencephalography (EEG) data categorization. Three secondary tables were created from the EEG data after they were transformed into spectrogram pictures. Deep learning is carried out once the acquired pictures are organized into folder structures using the spectrogram approach. For every job in the deep learning phase, 400 photos are acquired and fed into the GoogLeNet. Once deep learning was finished, the system that was provided was tested to see movements in the directions of up, down, left, and right in order to control the robot arm's movement. The robot arm is seen to carry out the intended motion over 90 % accuracy. To identify and enhance EEG signal states, a novel state identification approach is put forth in the work of Yang et al. [34]. To extract EEG characteristics along the time sequence, a Long Short-Term Memory Convolutional Neural Network (LSTM-CNN) was first created. Errors resulting from mental randomness or outside environmental influences may arise throughout this procedure. To address these mistakes, an actor-critic based decision making model was put out. Two networks make up the model, which may be used to forecast the ultimate signal state based on both the likelihood of the present signal state and the probabilities of previous signal states. A hybrid BCI real-time control system application is then suggested as a means of controlling a BCI robot. The idea of this work is to use EEG signals recorded from the brain using 5-channel Emotive insight BCI to control 6 DOF robotic arm using Arduino controller.

## Methodology

There are three layers beneath the skull that enclose and protect the brain. The pia mater, arachnoid mater and the Dura mater. The human brain can be classified into three main parts with respect to functionality. The stem or cerebellum, which is responsible for vital body function, the limbic system for emotions, and finally the cortex for processing and analyzing critical thinking. There are four regions in the cortex, the frontal, parietal, occipital and temporal lobes. Every one of these lobes is responsible for major functions such as critical thinking, memory storage, visual data processing, sensing and movement. According to the neurophysiology theory [35], [36], the nerve cells communicate between each other through synapses. The nervous system consists of two parts with respect to functions, the Central Nervous System CNS and Periphery Nervous System PNS. The CNS consists of Brain and Spinal cords, while the PNS consists of sensory neurons that connect sensory receptors to the body at different locations to the CNS [37], [38].

At the same time, nervous system cells have two categories, the neurons that have the ability to electrical signaling and the supporting glial cells that responsible for necessary function. A neuron is a specific type of cell that is different from normal cells. It has three parts, dendrites, cell body and axon. The dendrites have a high percentage of cytoskeletal protein and ribosomes that aid with information receiving and processing. The synapse consists of two parts, the presynaptic terminal and the postsynaptic specialization. In between these two parts, an extracellular space exists called the synaptic cleft that is used for communications between neurons via neurotransmitters. The axon is responsible for reading the integrated signals after it passes the cell body [39].

The axon length ranges from a few microns to up to one meter in the spinal cord, which is responsible for transmitting the signals to distant regions of the body. Therefore, in summary the dendrites are responsible for receiving information, the body for signal processing and the axon is responsible for the signal transferring. The basic function of the glial cells is to hold the neurons together. However, they perform an important function such as carrying on the ionic balance within the brain, aid with injury recovery and controlling the neurotransmitters in the synaptic clef. The CNS has four types of glial cells, astrocytes, oligodendrocytes, microglia, and ependymal cells. Each of these types performs an important function. Astrocytes are responsible for balancing the chemical environment for better neural signaling. The oligodendrocytes do the same function as the axon in CNS to aid faster signal transfer. Microglia assists with removing cellular debris from injury sites. Cerebrospinal fluid is produced by the ependymal cells. The electrochemical signaling inside the neurons is called the action potential AP [40].

The cause of AP is due to the ion exchange across the neuron membrane and the AP represents the temporary change in the membrane potential along the axon. The AP initiated in the cell body and travels in one direction. In this research work, the system is used for controlling a 6DOF robotic manipulator using Real-Time EEG data. The signals were collected non-invasively from the scalp. Then the signals are subjected to several steps such as preprocessing, artifact removal, feature extractions and classifications. Then sent to the Arduino controller to control the servomotors at each joint of the robotic arm.

The definition of EEG signal is that it is the measurement of the flowing current during synaptic excitations of the dendrites of neurons in the cerebral cortex [41], [42], [43], [44], [45].

The electrical dipoles between the body of the neuron and the dendrites create electrical potentials. The pumping of positive ions of Sodium $Na^+$, Potassium $K^+$, calcium $Ca^+$ and the negative of chlorine $Cl^-$ causes the current in the brain, and this current generate a magnetic field over the scalp that is measurable by EEG system.

There are five types of brain signals regarding the frequency ranges. These types are Gamma (>35Hz), Beta (12-35 Hz), Alpha (8-12Hz), Theta (4-8Hz) and Delta (0.5-4Hz). Each type of these signals is related to a specific Brain state as shown in Table 1.

**Table 1 Brain signal bands**

| No. | Brain Waves | Frequency Hz | Brain Location | Brain State |
|---|---|---|---|---|
| 1 | Delta | 0.5-4 | Frontal Lobe | Deep Sleep |
| 2 | Theta | 4-8 | Various depends on the | Light sleep, Meditation |
| 3 | Alpha | 8-12 | Occipital lobe | Relaxation |
| 4 | Beta | 13-35 | Distributed symmetrically | Active thinking |
| 5 | Gamma | >35 | Somatosensory cortex | High level cognitive |

From Table 1, like sleeping or dreaming each band of EEG is generated in a specific state of the mind. Such as sleeping or relaxing etc. as it explained below:

- Delta Waves: These signals are generated in deep sleep, and they are usually very slow waves.
- Theta Waves: These signals are dreamy state, they are generated in sleep and Meditation.
- Alpha Waves: These signals are generated in the rest state of the mind. The waves are noticeable when the eyes are closed.
- Beta Waves: These are generated when the person is active and conscious. The waves can be detected during tasks that require significant attention.
- Gamma Waves: The waves are used for advanced cognitive processing.

The brain signals collected using the Emotive insight have noises and errors [46]. Therefore, the signals need to be preprocessed in order to prepare them for feature extraction that helps with the classification. From Table 1 the Delta and Theta bands are more likely dominant during the unconscious state of the mind, while the Gamma band is dominant in the hyperactive state. For this reason, this work deals with Alpha and Beta for robotic applications.

The system component can be seen in Figure 1. The figure shows not only the hardware devices, but also the functional features of each part of the system.

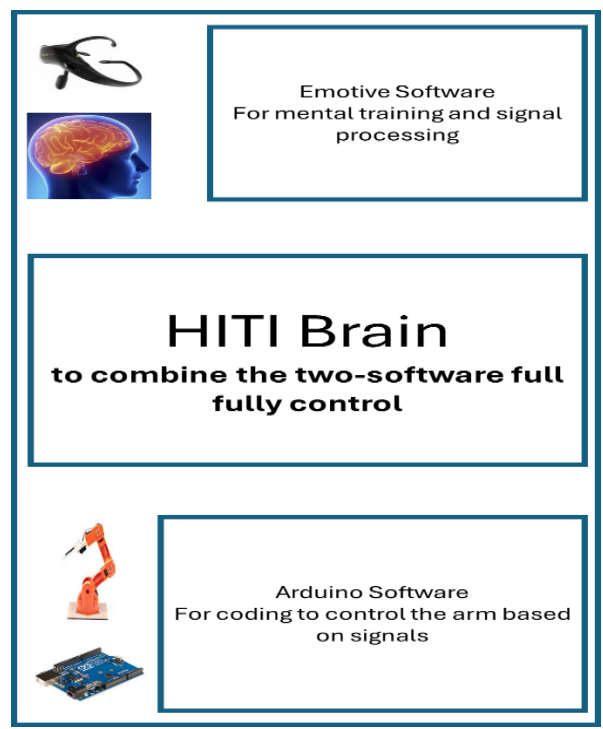

Figure 1 overall System component

*1- Brain computer Interface (BCI).*

BCI is a device that makes direct connections between electrical activity from the brain and an external device (Figure 2).

For recording the signals, there are typically three methods. The invasive, semi-invasive and non-invasive method [47].

Each method has its own pros and cons. The invasive method implants a micro electrode deep down the brain. However, the electrode kills every nerve it passes through until it reaches the required location. In this method, the signals are clean with less noise, high resolution and it can record data from a specific area of the brain. The semi-invasive method, medical implants are placed on the surface of the skull [48], [49].

The signals have higher resolution than the non-invasive method. In the non-invasive method, the sensors are placed over the head to record the electrical activities. It's the easiest method to record brain signals due to the easement of installations and it does not require any incisions or stiches. For this case the non-invasive method is used in this paper.

The device used in this work is Emotive Insight 2.0 with 5 channels. The device has 5 EEG sensors to detect brain activities and two reference (CML/DRL) sensors. One of the reference sensors is an active electrode (CML: Common Mode Sense) and the other is Passive electrode (DRL: Driven Right Leg).

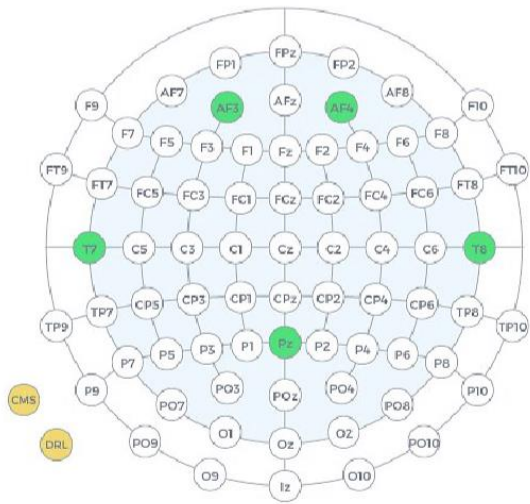

Figure 2 Electrode distribution of BCI on the scalp

As shown in Figure 2 the yellow circles represent the location of the reference sensors, and the green circle represents the sensor locations, which are the frontal, temporal and parietal part of the head. Emotive insight provides a non-invasive solution for detecting brain signals.

The device (as shown in Figure 3) detects a wide range of brain signals from three different parts of the brain, the Frontal lobe specifically from points (AF3) and (AF4), the Temporal lobe by points (T8) on the right and (T7) on the left and last the Parietal lobe from point (Pz). The device`s measurements are based on six keys: focus, Stress, Excitement, Relaxation, interest, and engagement. The Emotive Insight software is an open-source software that can be downloaded from the Emotive website.

There are several versions of the software, each one is used for different applications. Emotive BCI is the one that used in this work. The Emotive BCI is a desktop application. It can be used in both Mac and Windows.

The Emotive BCI allows the user to view and train the brain. The data streams for the Emotive BCI are classified in four categories: Mental Commands, Performance Metrics, Facial Commands and Motion sensors.

- *Mental Commands*: the software allows the user to train the brain based on commands such as Push, Pull, move right, Move left…. etc.

- *Facial Expression*: the software can trigger events based on facial expression. It records the brain signals based on smile, blink, wink …. etc.

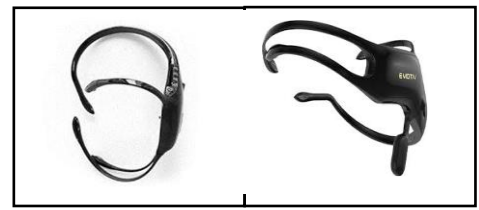

Figure 3 Emotive Insight BCI

The EEG signals are passing through a few stages of preprocessing. Firstly, the data is processed to remove sharp spikes. And then passed to a high pass filter to remove the DC offset and slow drift. The Emotive insight uses 2-second epochs and applies a Hanning-filter before performing the FFT. The POWER is calculated from the square of the amplitude in each frequency bin and output as $uV^{2}/Hz$.

The emotive software has the option of displaying all the signals detected from the brain. The brain data is recorded from five points (as the device used is emotive insight 5 channels). Figure 4 shows the Emotive software.

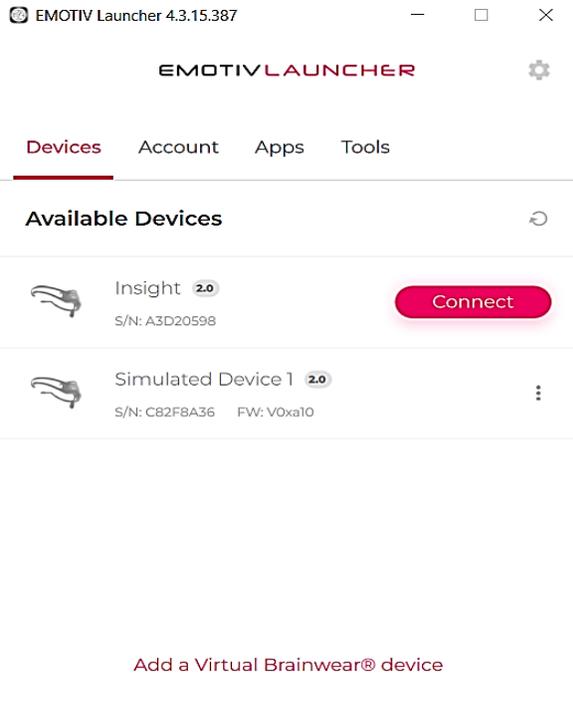

Figure 4 emotive software

The data sampling of Emotive Insight is 128 HZ with 16-bit resolution. The value of these data is affected by the different thoughts of the person.

The emotive software also has the ability to train the brain to do some mental commands like: Push, Pull, move left, right etc. as shown in Figure 5.

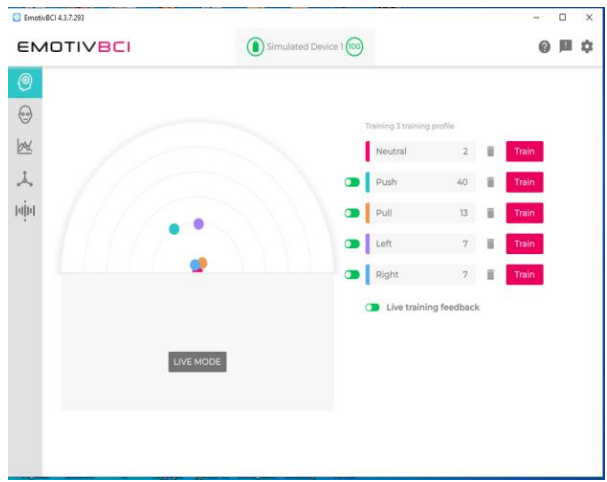

Figure 5 mental command

It also can record the signals of the brain due to facial expressions such as: smile, Blink, Wink left and right etc. (Figure 6). These records are useful to use for further applications in moving mobile robots[50], [51] or controlling multi degree of freedom robotic arms.

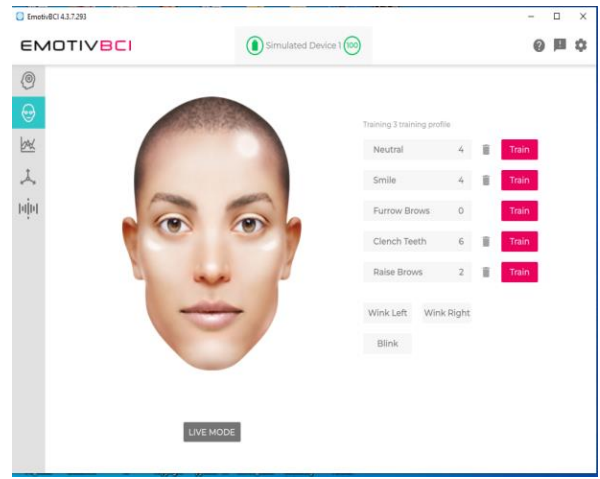

Figure 6 Facial expressions

### 2- 6 DOF Robotic Arm

The 6 DOF robotic arm has a wide range of applications in industry or for personal use as it is the closest to the human arm.

The Degree of Freedom (DOF) refers to the number of independent movements the robot arm can make.

A 3DOF robot arm can move up, down, right, left, in and out. The more Degree of freedom the arm has the wider range of motions it can perform.

The 6 DOF robotic arm can perform three linear motions (forward/back, right/left, up and down) and three rotational movements (tilting, twisting and rotating the arm itself).

This allows for a great precise and complex manipulation, mimicking the flexibility of the human arm.

One of the most important steps for controlling robot manipulator is to implement the complete and accurate system`s mathematical model.

The arm basically is a series of manipulators with revolute joints. The geometric configuration of the arm is made up of base, shoulder, elbow, and wrist. Figure (7) shows the robotic arm in correspondence to human arm.

Each joint except the wrist has a 1 DOF. The wrist has 2 DOF as it can move vertically and rotationally.

Each DOF in the arm is actuated by a servomotor. The end-effector is a two-finger gripper.

Each joint is connected to a servo motor that allows the joint to perform two movements. 6 servo motors lead to 12 movements for the arm.

The combination of brain thoughts enables the arm to perform complex motions such as Pick and Place.

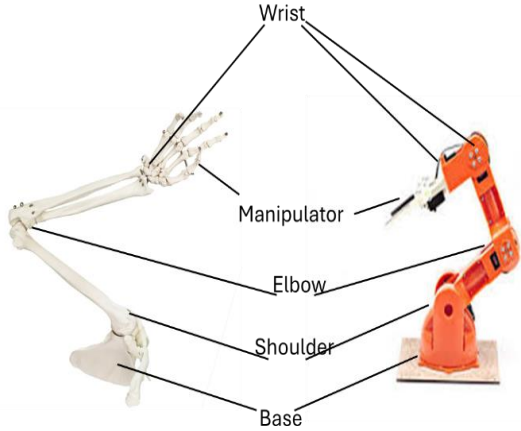

Figure 7 Robotic/Human arm

- *Forward kinematic Model.*

The kinematic model in this paper is achieved using Denavit-Hartenberg (DH) parameter. Figure 8 presents the kinematic model of the robotic arm in L position.

The Base, shoulder and elbow are moving the tool point to its desired position, the orientation of the end-effector is implemented by the wrist joints.

DH works with twist angle $\alpha_{i-1}$, link length $a_{i-1}$, link offset $d_i$ and joint angle $\theta_i \{ \alpha_{i-1}, a_{i-1}, d_i, \theta_i \}$.

As shown in Figure 9 a coordinate system is attached to each link of the manipulator. Table 2 lists the DH parameter of the robotic arm [12].

The derivation of the links (joint expressed as $i$ in its previous neighboring joint $i-1$ was derived and presented in Equation (1) which represents the overall matrix of the end-effector to the base of the robotic arm.

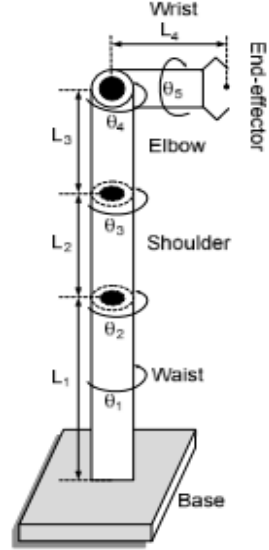

Figure 8 kinematic model

**Table 2 DH parameter of Robotic arm**

| Symbols | Joints | | | | | |
|---|---|---|---|---|---|---|
| | 1 | 2 | 3 | 4 | 5 | 6 |
| $\alpha_{i-1}$ | 0 | -90 | 0 | 0 | -90 | 0 |
| $a_{i-1}$ | 0 | 0 | $L_2$ | $L_3$ | 0 | 0 |
| $d_i$ | $L_1$ | 0 | 0 | 0 | 0 | $L_4$ |
| $\theta_i$ | $\theta_1$ | $\theta_2 - 90$ | $\theta_3$ | $\theta_4$ | $\theta_5$ | 0 |

$$^0_6T = \begin{bmatrix} C_1C_5S_{234} + S_1S_5 & -C_1S_{234}S_5 + C_1C_{234} & C_1C_{234} & C_1A \\ -S_1C_5C_{234} - C_1S_5 & S_1C_{234}C_5 + C_1C_5 & S_1C_{234} & S_1A \\ C_{234}C_5 & -C_{234}S_5 & -S_{234}B & S_1A \\ 0 & 0 & 0 & 1 \end{bmatrix}$$
(1)

where

$$A = L_2S_2 + L_3S_{23} + L_4C_{234}$$
$$B = L_1 + L_2C_2 + L_3C_{23} - L_4S_{234}$$

From equation (1), the 3*3 matrix (the first three rows and first three columns) is the rotation matrix. The last column represents the position (*x, y, z*) of the end-effector with respect to the base.

- Inverse kinematic Model.

In practical robotic systems, inverse kinematics have more potential applications. With inverse kinematics, the computation of the joint angle required for achieving the required position and orientation can be calculated [52].

In inverse kinematic, the joint angles ($\theta_1, \theta_2, \theta_3, \theta_4$) must be calculated, while $\theta_5$ is directly given by the desired orientation for object manipulation [53], [54].

The transformation matrix from the tool to base is given by:

$$^{Base}_{Tool}T = \begin{bmatrix} n_x & o_x & a_x & p_x \\ n_y & o_y & a_y & p_y \\ n_z & o_z & a_z & p_z \\ 0 & 0 & 0 & 1 \end{bmatrix}. \quad (2)$$

The first 3*3 matrix represents the rotation, and the last column represents the translation of the end-effector with respect to the base.

$$\theta_1 = Atan2(p_x, p_y) \quad (3)$$
$$s_{234} = c_1 a_x + s_1 a_y \quad (4)$$
$$c_{234} = a_z \quad (5)$$
$$\theta_{234} = Atan2(s_{234}, c_{234}) \quad (6)$$
$$c_3 = \frac{(c_1 p_x + s_1 p_y + l_4 s_{234})^2 + (p_z - l_1 + l_4 c_{234})^2 - l_2^2 - l_3^2}{2 l_2 l_3} \quad (7)$$
$$s_3 = \mp\sqrt{1 - c_3^2} \quad (8)$$
$$\theta_3 = Atan2(s_3, c_3) \quad (9)$$
$$c_2 = \frac{(c_1 p_x + s_1 p_y + l_4 s_{234})(c_3 l_3 + l_2) - (p_z - l_1 + l_4 c_{234}) s_3 l_3}{(c_3 l_3 + l_2)^2 + s_3^2 l_3^2} \quad (10)$$
$$s_2 = \frac{(c_1 p_x + s_1 p_y + l_4 s_{234})(s_3 l_3) - (p_z - l_1 + l_4 c_{234})(c_3 l_3 + l_2)}{(c_3 l_3 + l_2)^2 + s_3^2 l_3^2} \quad (11)$$
$$\theta_3 = Atan2(s_2, c_2) \quad (12)$$
$$\theta_4 = \theta_{234} - \theta_2 - \theta_3 \quad (13)$$

The experiment implemented in two conditions, the first one is quiet environment and the second is noisy environment (such as two people were talking next to the subject).

The system consists of three different software working together.

The Emotive software (for Emotive BCI), the Arduino UNO software and the HITI brain software that combines the two previous software together.

The HITI brain is software that can make communication between Arduino board and Emotive EEG headset very easy.

It allows the Arduino to receive mental commands from Emotive insight headset. Figure 9 shows the system block diagram.

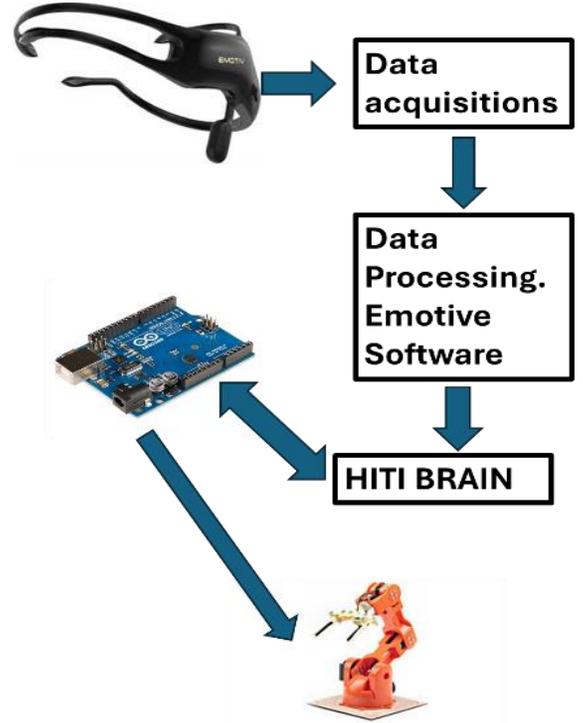

Figure 9 the system block diagram

The target movement in this work is as follows:

1- Test all the movement using the mental commands separately.
2- Make the full pick and place by combining different mental commands.

As mentioned earlier, the 6 DOF robotic arm is very close to the human arm. The work of the robotic arm is more important than the outside shape of the robotic arm.

Therefore, the main aim in this work is the actuators that move the robot links together (aka. The servo motors).

The Arduino code defines the motors work and posterization of which motor work is defined in the code according to the mental commands.

Figure 10 shows the flowchart of the system working. The actuators' movement is based on the predicted category.

The three-software cooperating between each other in order to make a language that the robots understand.

The BCI read the signals using the electrodes connected to the cortex, then transfer the signal to the Emotive software for recording and further process for noise reduction and classification.

The Arduino UNO has the codes for controlling the actuator and implementing the path in the robotic arm workspace.

The HITI brain software connects the Arduino code and the Emotive software, executes the code with the database from the emotive software, and transforms it into actual motion to the connected arm. The software working principle is building a GUI to control the actuators according to the code written in Arduino IDE.

It's important to mention that the subject needs to be trained before the final test. The emotive software has a training feature through a designed experiment.

The subject sits in a relaxed position in front of the screen. a cube in the center of the screen appears and the subject needs to imagine moving it forward, backward, right, left, up and down.

This experiment is important for two reasons. First, is to sharpen the brain to produce the same pattern of signals for each movement.

Secondly, store these signals for referencing and comparing them with the real time signals in the final test, so the HITI brain can command the Robot arm to move according to the mental command by the help of the Arduino code.

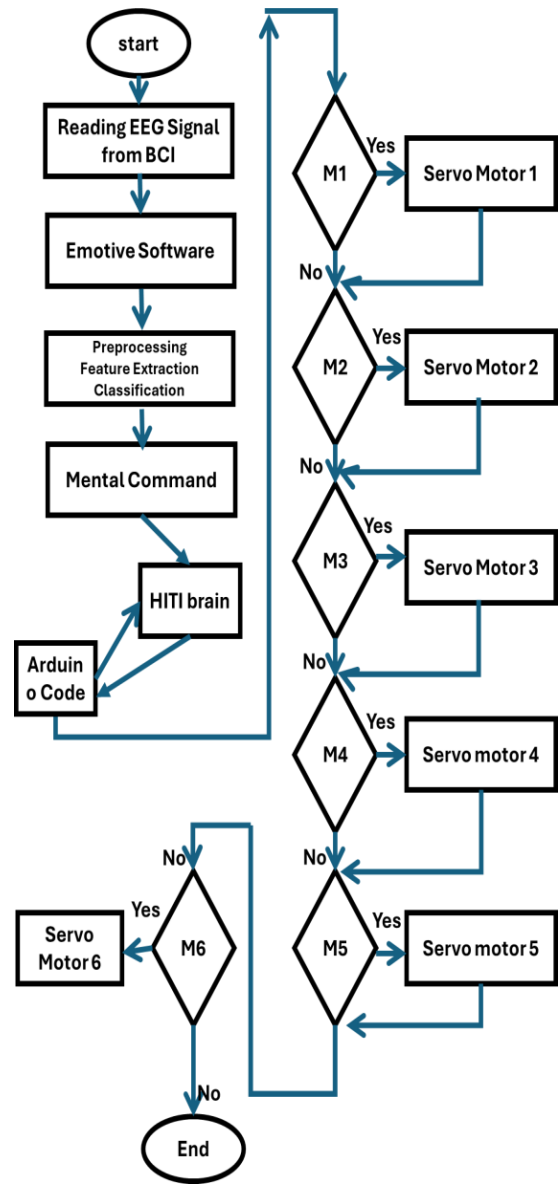

Figure 10 The system flowchart

The servo motor is controlled by a pulse width modulated signal. The Arduino send a pulse with 1ms and 2 ms long every 20ms to the servo motor. Figure 11 shows the servo signal.

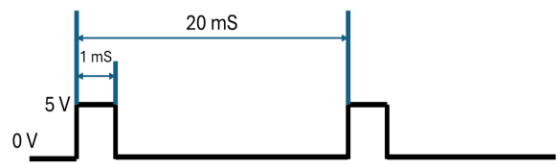

Figure 11 Servo signal

The servo motor rotates the shafts to the central position during the 1 ms pulse. Several pulses rotate the motor enough to move the links to the desired location. The servo library in Arduino generates the necessary PWM signals to send it to the servo.

During Motor imagery (imagining movement without actual physical motion), a distinct pattern is shown in the EEG signals in the Alpha and Beta bands. These patterns play a great role in decoding mental commands to control the robotic arm or other devices. In the state of motor imagery task, the Alpha power tends to decrease in the sensorimotor cortex. This is called Event-Related Desynchronization (ERD). The Brain tends to reduce the Alpha-band to prepare for motor actions. On the contrary, during motor imagery, the Beta wave is increasing. This increase is called Event-Relate Synchronization (ERS). The ERS occur as a response to imagining moving a part of the body. Before the controlling of the robotic arm, the Emotive EEG signals have to be passed through several steps, the preprocessing, Feature Extraction and classification.

- A- *Preprocessing*
  As mentioned before, the EEG signals contain noises and artifacts that are caused by Eye blinking, moving the face muscles. Using band pass filter is essential to remove all other bands and focus only on Alpha (8-12 Hz) and Beta (13-30 Hz). As for the artifacts, the Emotive insight automatically removes them.
- B- *Feature Extraction*
  In this work, Wavelet transform, the same method in [47], [49] was used for feature extraction. To capture the Alpha and Beta band waves, 4-level decomposition of the EEG signals must be executed. The common features captured were as follows:
  - Energy of wavelet coefficient: the energy is extracted using the equation
    $E = \sum_{i-1}^{N} |D_i|^2$ ----(14)
    Where $D_i$ is the detailed coefficient corresponding to Level N.
  - Mean: the average of signals in the band.
    $\mu = \frac{1}{N}\sum_{i=1}^{N}(D_i - \mu)^2$ -----(15)
  - Other features such as variance, skewness, kurtosis, max and min.

- C- *Classification*
  This stage comes after the extractions of features of the EEG signals that accomplished in the previous step. For each segment of the EEG, the feature vector needs to be labeled for mental commands (i.e. Push=1, Pull=2, etc.) to train the classifier to detect them and use it to control the robotic arm.

## Experimental Results

The previous section explained in short description how the system works. First is the trial stage. In this stage the subject is in need of training his brain via a simple experiment.

The main aim of this is to make the subject brain get used to the device and second to make the brain produce the same patterns of signals for a specific action.

For example, the first training is for push motor imagery, when the subject is imagining pushing an object. After several trials the brain will produce the same pattern of signals for such action.

The Emotive software comes with experiment features for subject training as shown in Figure 12. From the figure we can see the subject sitting Infront of the screen where there is a cube in the middle.

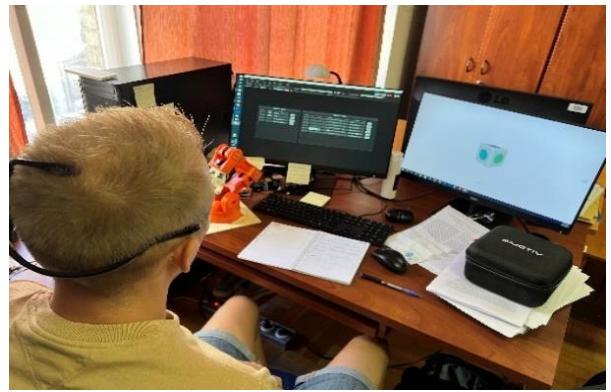

Figure 12 Subject Training

The subject has to relax and focus on the cube to perform several actions, such as Push, Pull, Lift, Drop, move right and move left.

All the signals from the different trainings will be stored and preprocessed for artifact removal, feature extractions and classifications.

Figure 13 shows the features (Energy and mean) extracted from Beta and Alpha waves for four mental commands.

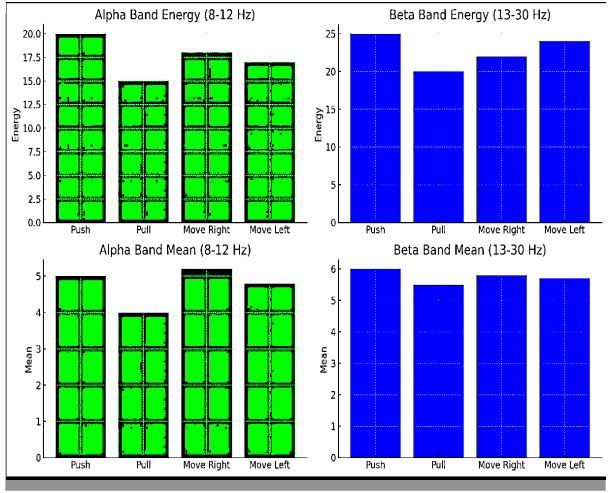

Figure 13 Mean and Energy for EEG signal

The Arduino code uses the stored signals as a reference to convert them to commands to control the robotic arm. The result of this work is divided into two parts:

1- Separate movement.
2- Full pick and place movement.

Both of the parts were executed in two conditions, the Quiet mode, where there is no extra noise in the environment.

The second condition is where there is noise in the environment, as people talking or playing music next to the subject.

*1- Separate movements (Mental Command)*
   Each joint in the robotic arm is connected to a servomotor. The servomotor moves in two directions. That means each movement needs 1 mental command. For example, mental command move right will result in rotating the servo 90 degrees CW. And the mental command move left will force the motor to rotate 90 degrees CCW as shown in Figure 14 below. The movement of the Base joint was chosen for this section which also is shown in Figure 15 a,b,c.

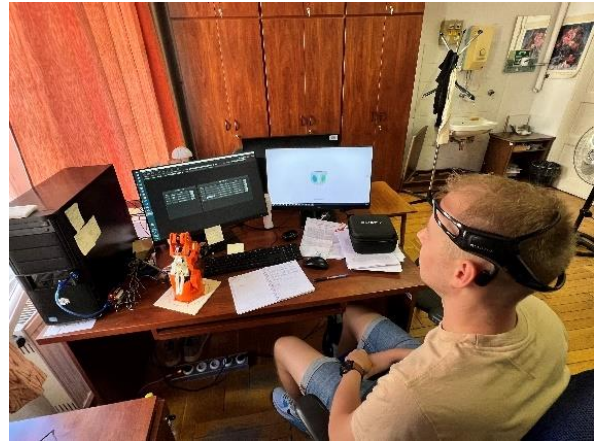

Figure 14 Mental Command

As it shown in Figure 15, the initial position of the robot arm was in figure b at 0 degree the figure a is moving right CW and figure c is rotating Left CCW. Beta waves are related with active thinking, problem solving, concentration, and increased attentiveness.

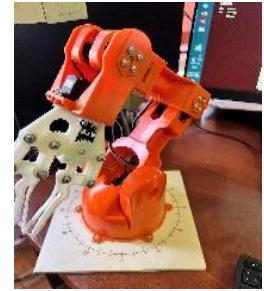

a

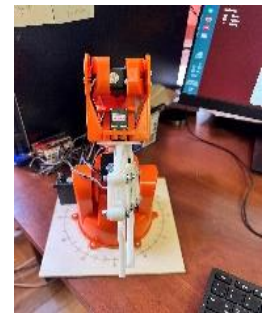

b

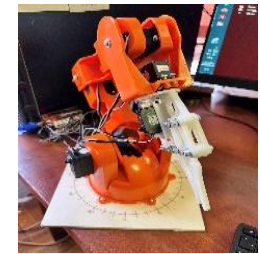

c

Figure 15 Robot arm (Base) movement

In an agitated environment, beta waves rise. This is a method of being attentive in the face of imagined threats from ambient noise.

Figures 16 and 17 show the beta waves amplitude changing as Figure 16 represents the beta waves in a quiet environment the low beta waves are increasing.

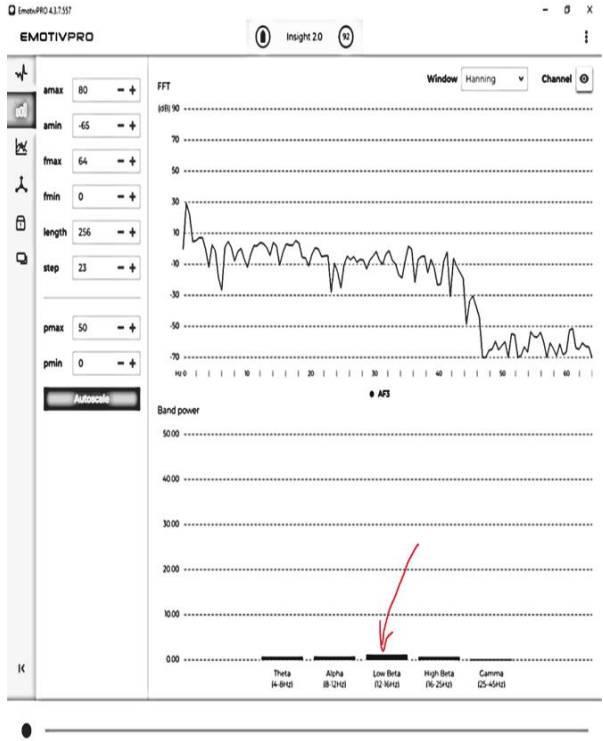

Figure 16 Low Beta waves in a quiet environment

Figure 17 shows the High beta waves in a noisy environment a music was playing next to the subject.

From Figure 17, high beta waves were arising as the subject was in a high focus as the noise environment affected the training. The same results were achieved for the Robotic arm movement.

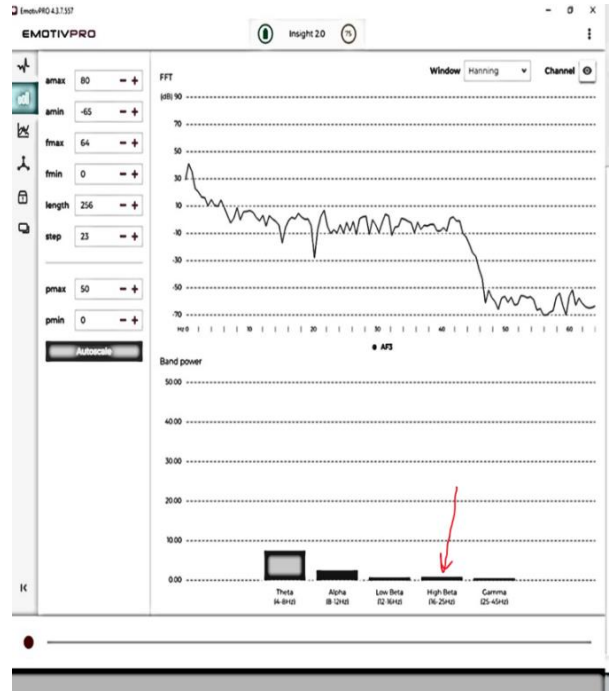

Figure 17 High Beta waves in a noisy environment

2- *Separate movement (Facial Expression).*
The BCI records the signals of the brain based on special facial actions. Every time a person winks, blinks, smiles or does special action in the face, the brain produce different type of signals.
Moreover, since the servomotor rotates in two directions, each direction requires one facial action.
For example, a smile will force the motor to rotate CW. In addition, the raised eyebrows facial action leads to rotating the motor CCW. Figure 1 a, b shows the facial expression.

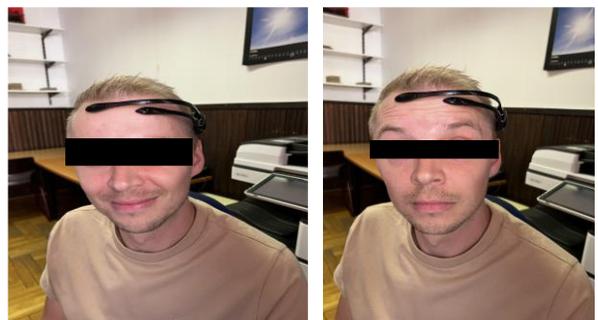

a            b
Figure 18 Facial expression

Figure 19 a, b, c shows movement of the elbow joint that was chosen for this section.

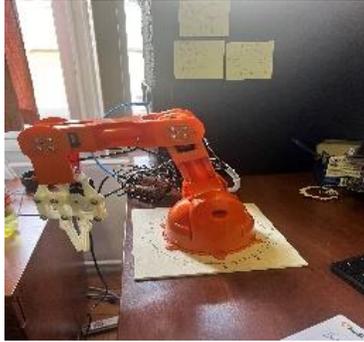

a

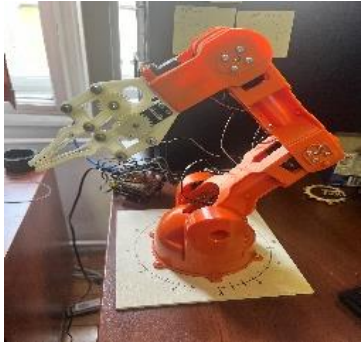

b

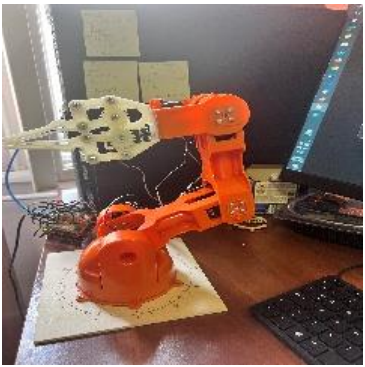

c

Figure 19 Elbow movement

From the Figures above, Figure 'b' represents the zero position, Figure 'a' is for CCW shoulder joint and Figure 'c' shows the CW rotation of the shoulder joint.

3- *Full Pick and Place movement.*

This is a little hard to perform since the work is limited by the actions available in the emotive software.

The robotic arm has 6 joints and each joint moves in two directions. So, the total brain actions needed for the full arm are 12 actions.

Therefore, in order to perform this complex movement with the limited resources in hands, the combinations of mental commands and facial expression were implemented in this section. Table 3 presents each joint with their movement and which action is connected to it.

**Table 3 joints and brain actions.**

| Joint | | CW rotation | CCW rotation | Mental Command | Facial Expression |
|---|---|---|---|---|---|
| Base | | Move Right | Move Left | **Yes** | No |
| Shoulder | | Push | Pull | **Yes** | No |
| Elbow | | Lift | Drop | **Yes** | No |
| Wrist | Rotational | Raise brows | Furrow brows | No | **Yes** |
| Wrist | Transitional | Wink left | Wink Right | No | **Yes** |
| Manipulator | | Smile | Clench Teeth | No | **Yes** |

Figure 20 shows how full movement of an object is performed from one point to another.

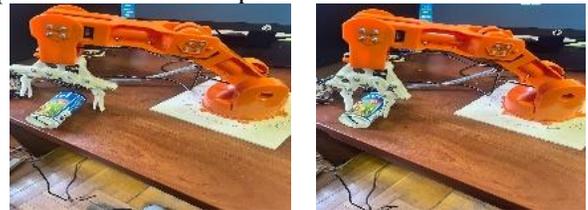

a        b

Figure 20 Pick and Place

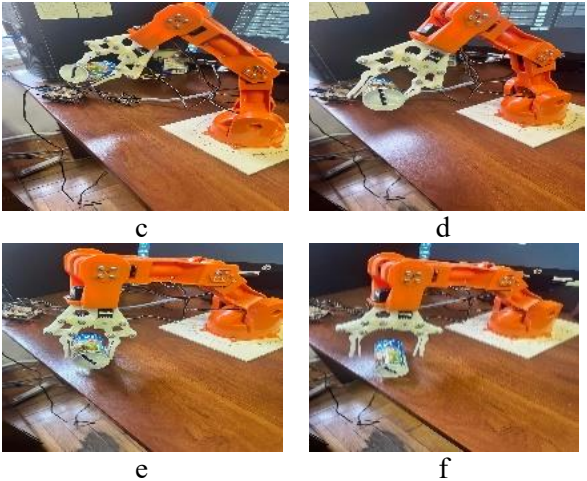

Figure 20 Pick and Place (Cont'd).

4- *Safety and Security.*

The system has two safety conditions in order to avoid colliding with other objects on the way and to avoid if the signal is too powerful that makes the joint move more than the desired motion. The first solution is to set the minimum and maximum interval range for the servo. This can be done in the code. The second solution is in the HITI brain software (Figure 21). When linking the joint movement to brain signals, there is an option called a threshold. This can determine the power the signal should reach before moving the motor. The safety of this system is very crucial to ensure smooth and safe working of the arm, as the system is also applicable for a prosthetic arm with 6 joints. And since the prosthetic attached to the body is different than the robotic arm, putting the safety conditions is required always.

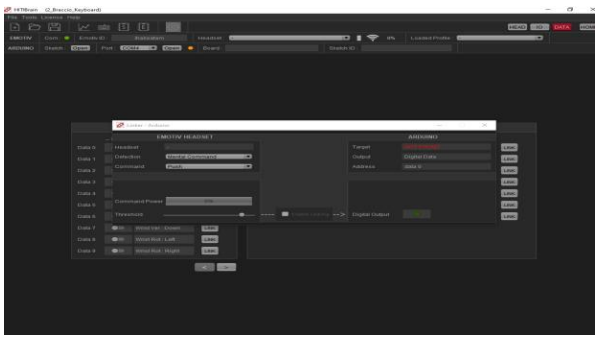

Figure 21 HITI software safety option

## Conclusion

Disability problem is widespread in the last decade. Paralyzed or people with lost limbs (especially the upper limb) face a real challenge performing daily life activities.

The aim of this research is to help them to perform simple daily life activities using their brain signal.

The work shows great results regarding the motion control of the robotics arm, and at the same time the system is safe and smooth.

The patient before using the system needs to be trained for a period of time, and this time is different from person to person depends on power of focus in the patient. The overall system can be applicable for prosthetics and other actuators such as bikes, mobile robots and cars.